\documentclass[10pt,twocolumn,letterpaper]{article}

\usepackage{wacv}
\usepackage{times}
\usepackage{epsfig}
\usepackage{graphicx}
\usepackage{amsmath}
\usepackage{amssymb}
\usepackage[accsupp]{axessibility}  % Improves PDF readability for those with disabilities.
\usepackage{pifont}
\usepackage{subfig}
\usepackage{multirow}

\def\wacvPaperID{****} % Enter the WACV Paper ID here

\wacvfinalcopy % *** Uncomment this line for the final submission

\ifwacvfinal
\fi

\ifwacvfinal
\usepackage[breaklinks=true,bookmarks=false]{hyperref}
\else
\usepackage[pagebackref=true,breaklinks=true,colorlinks,bookmarks=false]{hyperref}
\fi

\begin{document}

%%%%%%%%% TITLE
\title{Robust Lane Detection via Expanded Self Attention}

\author{
	Minhyeok Lee \quad
	Junhyeop Lee \quad
	Dogyoon Lee \quad
	Woojin Kim \quad
	Sangwon Hwang \\
	Sangyoun Lee$^{*}$ \quad
	%\and 
	\vspace{0.01cm}\\
	Yonsei University School of Electrical and Electronic Engineering \\
	{\tt\small \{hydragon516,jun.lee,nemotio,woojinkim0207,sangwon1042,syleee\}@yonsei.ac.kr}
}

\maketitle

\ifwacvfinal
\thispagestyle{empty}
\fi

%%%%%%%%% ABSTRACT
\begin{abstract}
   The image-based lane detection algorithm is one of the key technologies in autonomous vehicles. Modern deep learning methods achieve high performance in lane detection, but it is still difficult to accurately detect lanes in challenging situations such as congested roads and extreme lighting conditions. To be robust on these challenging situations, it is important to extract global contextual information even from limited visual cues. In this paper, we propose a simple but powerful self-attention mechanism optimized for lane detection called the Expanded Self Attention (ESA) module. Inspired by the simple geometric structure of lanes, the proposed method predicts the confidence of a lane along the vertical and horizontal directions in an image. The prediction of the confidence enables estimating occluded locations by extracting global contextual information. ESA module can be easily implemented and applied to any encoder-decoder-based model without increasing the inference time. The performance of our method is evaluated on three popular lane detection benchmarks (TuSimple, CULane and BDD100K). We achieve state-of-the-art performance in CULane and BDD100K and distinct improvement on TuSimple dataset. The experimental results show that our approach is robust to occlusion and extreme lighting conditions. 
\end{abstract}

%%%%%%%%% BODY TEXT
\section{Introduction}
Advanced Driver Assistance Systems (ADAS), which are a key technology for autonomous driving, assists drivers in a variety of driving scenarios owing to deep learning. For ADAS, lane detection is an essential technology for vehicles to stably follow lanes. However, lane detection tasks, which rely on visual cues such as cameras, remain challenging owing to severe occlusions, extreme changes in the lighting conditions, and poor pavement conditions. Even in such difficult driving scenarios, humans can sensibly determine the positions of lanes by recognizing the positional relationship between the vehicles and surrounding environment. This remains a difficult task in image-based deep learning.

\begin{figure}
	\setlength{\belowcaptionskip}{-24pt}
	\begin{center}
		\includegraphics[width=\linewidth]{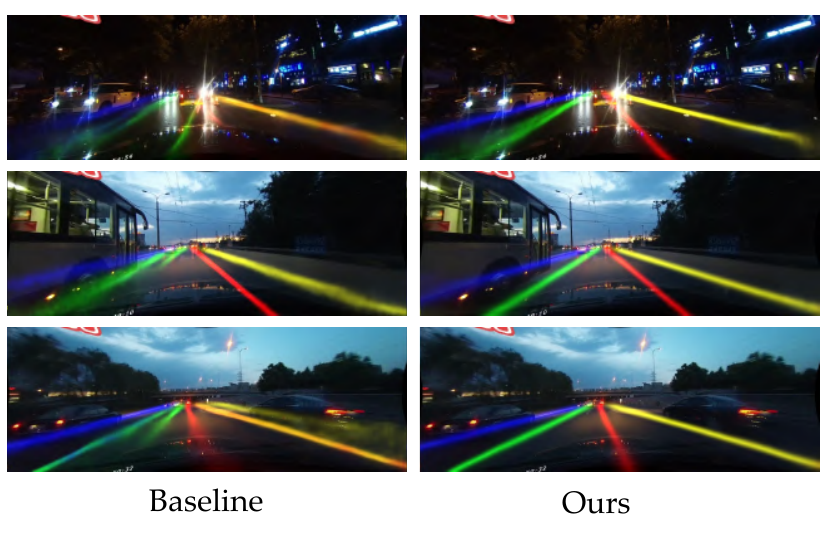}
		\\[-2ex]
		\caption{Compare our method with the baseline model. Our approach shows robustness in a variety of occlusion and low-light conditions.}
		\label{fig:intro}
	\end{center}
\end{figure}

The most widely used lane detection approach in image-based deep learning is segmentation-based lane detection~\cite{neven2018towards, pan2017spatial, hou2019learning, ghafoorian2018gan, lee2017vpgnet, liu2020lane, chang2019multi, lo2019multi, chen2018efficient}. These works learn in an end-to-end manner whether each pixel of the image represents the lane. However, it is very difficult to segment lane areas that are not visible by occlusion. To solve this problem, the network must capture the scene context with sparse supervision. Therefore, some works~\cite{pan2017spatial, hou2019learning} also introduce message passing or attention distillation. In~\cite{ghafoorian2018gan}, adversarial learning was applied to generate lanes similar to the real ones. These approaches can capture sparse supervision or sharpen blurry lanes. However, segmenting every pixel to detect lanes can be computationally inefficient.

To simplify the lane detection process and increase efficiency, some works \cite{qin2020ultra, yoo2020end, chougule2018reliable} consider the problem of lane detection a relatively simple task and adopt the classification method. In~\cite{qin2020ultra}, a very fast speed was achieved by dividing the image into a grid of a certain size and determining the position of the lane with row-wise classification. However, these methods do not represent lanes accurately, nor do they detect relatively large numbers of lanes.

To address the shortcomings of the semantic segmentation and classification methods described earlier, we propose a novel self-attention module called the Expanded Self Attention (ESA) module. Our modules are designed for segmentation-based lane detection and can be attached to any encoder-decoder-based model. Moreover, our method does not increase the inference time because the ESA module is removed in the testing phase. To make the model robust to occlusion and difficult lighting conditions, ESA module aims to extract important global contextual information by predicting the occluded location in the image. Inspired by the simple geometry of lanes, ESA modules are divided into HESA (Horizontal Expanded Self Attention) and VESA (Vertical Expanded Self Attention). HESA and VESA extract the location of the occlusion by predicting the confidence of the lane along the vertical and horizontal directions, respectively. Since we do not provide additional supervisory signals for occlusion, predicting occlusion location by the ESA module is a powerful help for the model to extract global contextual information. Details of the ESA module will be presented in Section~\ref{ESA}.

Our method is tested on three popular datasets (TuSimple, CULane and BDD100K) containing a variety of challenging driving scenarios. Our approach achieves state-of-the-art performance in the CULane and BDD100K datasets, especially in CULane, surpassing the previous methods with a F1 score of 74.2. We confirm the effectiveness of the ESA module in various comparative experiments and demonstrate that our method is robust under occlusion and extreme lighting conditions. In particular, the results in Figure~\ref{fig:intro} show that our module shows impressive lane detection performance in various challenging driving scenarios. 

Our main contributions can be summarized as follows:

\begin{itemize}
	\item We propose a new Expanded Self Attention (ESA) module. The ESA module remarkably improves the segmentation-based lane detection performance by extracting global contextual information. Our module can be attached to any encoder-decoder-based model and does not increase inference time.
	
	\item Inspired by the simple lane geometry, we divide the ESA module into HESA and VESA. Each module extracts the occlusion position by predicting the lane confidence along the vertical and horizontal directions. This makes the model robust in challenging driving scenarios.
	
	\item The proposed network achieves state-of-the-art performance for the CULane \cite{pan2017spatial} and BDD100K \cite{yu2018bdd100k} datasets and outstanding performance gains under low-light conditions.
\end{itemize}

\begin{figure*}
	\setlength{\belowcaptionskip}{-10pt}
	\begin{center}
		\includegraphics[width=1\linewidth]{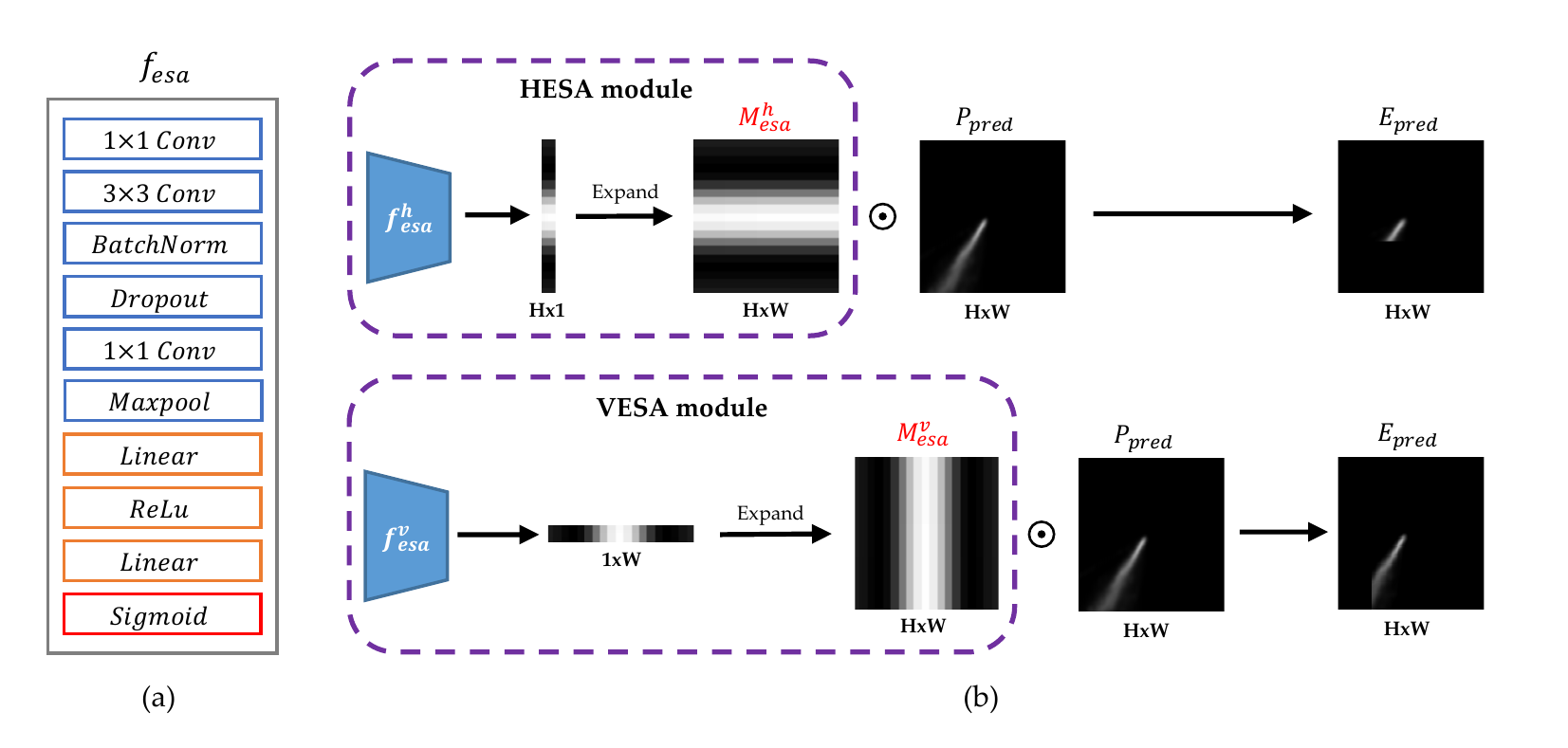}
	\end{center}
	\vspace{-7.5mm}
	\caption{(a) Structure of ESA encoder $f_{esa}$. (b) Details of the Horizontal Expanded Self Attention (HESA) module (top) and Vertical Expanded Self Attention (VESA) module (bottom). The only difference between the two modules is the expansion direction of the ESA encoder output. Operator $\odot$ is defined as an element-wise product.}
	\label{fig:esa}
\end{figure*}

\section{Related Work}
\noindent
\textbf{Lane Detection.} The use of deep learning for lane detection has been increasingly popular. Owing to the success of deep learning in the computer vision field, many studies have been proposed by adopting deep learning technique on lane detection for advanced driving assistant system, particularly for autonomous driving~\cite{neven2018towards, pan2017spatial, hou2019learning, qin2020ultra, yoo2020end}. This approach performs better than hand-crafted methods~\cite{ek2004lane, sun2006hsi, wang2000lane, kim2008robust}. There are two main deep-learning-based approaches: 1) classification-based and 2) segmentation-based approaches.

The first approach considers lane detection a classification task \cite{qin2020ultra, yoo2020end, chougule2018reliable}. Some works \cite{qin2020ultra, yoo2020end} applied row-wise classification for the detection of lanes, thereby excluding unnecessary post-processing. In particular, \cite{qin2020ultra} achieved high-speed performance by lightening the model. However, in the classification method, the performance depends on how many times the position of the lane is subdivided. In addition, it is difficult to determine the shape of the lane accurately.

Another approach to lane detection is to consider it a semantic segmentation task \cite{neven2018towards, pan2017spatial, hou2019learning, hou2020inter, lee2017vpgnet}. Neven~\etal~\cite{neven2018towards} performs instance segmentation by applying a clustering method to line mark segmentation. Moreover, Lee~\etal~\cite{lee2017vpgnet} proposes multi-task learning that simultaneously performs grid regression, object detection, and multi-label classification guided by the vanishing point. Multi-task learning provide additional supervisory signals. However, the additional annotations required for multi-task learning are expensive. Pan~\etal~\cite{pan2017spatial} applies a message passing mechanism between adjacent pixels. This method overcomes lane occlusion caused by vehicles and obstacles on the road and recognizes lanes in low-light environments. However, this message passing method requires considerable computational cost. To solve the slow speed of the method in~\cite{pan2017spatial}, Hou~\etal~\cite{hou2019learning} proposes the Self Attention Distillation (SAD) module and achieve a significant improvement without additional supervision or labeling while maintaining the number of parameters in the model. However, in the SAD module, knowledge distillation is conducted from deep to shallow layers, which only enhances the inter-layer information flow for the lane area and does not provide an additional supervisory signal for occlusion. Our work is similar to \cite{hou2019learning}, in that it uses the self-attention module. However, it adopts a new self-attention approach in a completely different way. To overcome occlusion problems, the proposed ESA module calculates the confidence of the lane that is deeply related to the occlusion. By using lane confidence, the model can reinforce the learning performance for these areas by providing a new supervisory signal for occlusion.

\noindent
\textbf{Self-attention.} Self-attention has provided significant improvements in machine translation and natural language processing. Recently, self-attention mechanisms are used in various computer vision fields. The non-local block \cite{wang2018non} learns the relationship between pixels at different locations. For instance, Zhang \etal~\cite{zhang2019self} introduces a better image generator with non-local operations, and Fu \etal~\cite{fu2019dual} improves the semantic segmentation performance using two types of non-local blocks. In addition, self-attention can emphasize important spatial information of feature maps. \cite{park2018bam, woo2018cbam} showed meaningful performance improvement in classification by adding channel attention and spatial attention mechanisms to the model.

The proposed ESA module operates in a different way than the previously presented module. The ESA module extracts the global context of congested roads to predict areas with high lane uncertainty and to emphasize those lanes.

\section{Proposed Approach}
\subsection{Overview}
Unlike general semantic segmentation, lane segmentation conducts segmentation by predicting the area in which the lane is covered by objects. Therefore, lane segmentation tasks must extract global contextual information and consider the relationship between distant pixels. In fact, self-attention modules with non-local operation \cite{wang2018non} can be an appropriate solution. Several works \cite{zhu2019asymmetric, fu2019dual, huang2019ccnet} prove that non-local operations are effective in semantic partitioning where global contextual information is important. However, in contrast to the complex shape in general semantic segmentation, the lane has a relatively simple geometric shape in lane segmentation. This makes non-local operations inefficient.

If the network can extract occluded locations, lanes that are invisible owing to occlusions are easier to segment. The location information of occlusions becomes more important than their shape owing to the simple lane shape. Therefore, rather than extracting the high-level occlusion shape, it is more effective to extract the low-level occlusion position. By using this positional information, the ESA module can extract the column or row-wise confidence of lanes by itself. The confidence indicates that the model knows the location of the occlusion based on the global contextual information of the scene.

\begin{figure}
	\setlength{\belowcaptionskip}{-24pt}
	\begin{center}
		\includegraphics[width=0.9\linewidth]{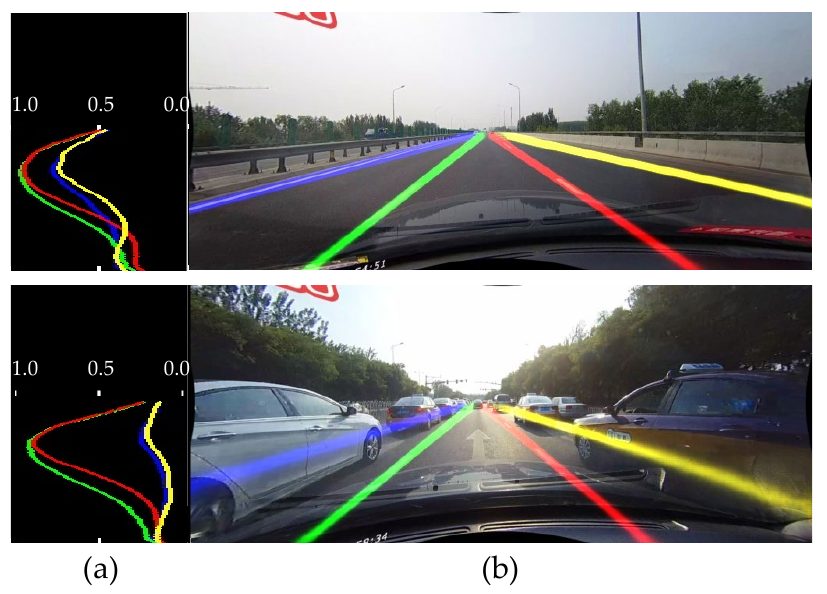}
		\caption{(a) Graph of ESA encoder $f^{h}_{esa}$ output and (b) predicted lane probability map. The output of the ESA encoder represents the lane confidence of each row. The graph and lane are matched with the same color, and the graph shows only the area in which the lane exists.}
		\label{fig:conf}
	\end{center}
\end{figure}

\subsection{Expanded Self Attention}
\label{ESA}
The ESA module aims to extract global contextual information by recognizing the occluded area. The structure of the ESA module is inspired by the fact that the lane is a line that spreads from the vanishing point. Due to the simple shape of the lane, it is efficient to predict the confidence along the vertical or horizontal direction of the lane in order to estimate the location of the occlusion. Therefore, we divide the ESA module into HESA and VESA according to the direction to extract the lane confidence. Furthermore, all ESA modules are only used in the training phase and are removed in the testing phase. Therefore, our method has the same inference time and number of parameters as the baseline model.

\begin{figure*}
	\setlength{\belowcaptionskip}{-10pt}
	\begin{center}
		\includegraphics[width=0.9\linewidth]{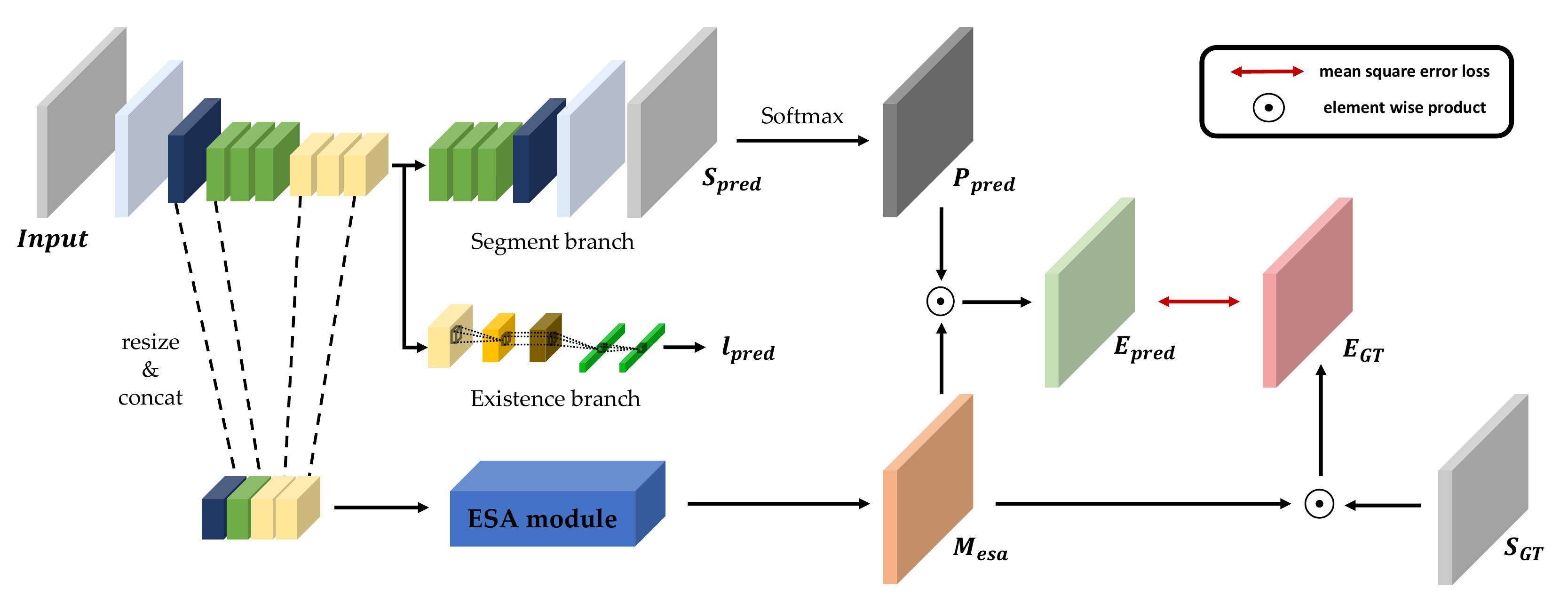}
	\end{center}
	\vspace{-7.5mm}
	\caption{The neural network architecture. The model is a combination of the existence branch and ESA module in the baseline model. The existence branch outputs the probability of existence of each lane and the ESA module generates an ESA matrix.}
	\label{fig:network}
\end{figure*}

Figure~\ref{fig:esa} shows two types of ESA modules, HESA and VESA. Both modules have an ESA encoder $f_{esa}$ consisting of convolution layers and fully connected layers. The ESA encoders of the HESA and VESA modules are defined as $f^{h}_{esa}$ and $f^{v}_{esa}$, respectively. The only difference between the two encoders is the length of the output vector. For the HESA modules, the output shape of $f^{h}_{esa}$ is $\mathbb{R}^{C\times H\times 1}$, where $C$ is the maximum number of lanes, and $H$ is the height of the image. This output will be expanded horizontally and will be equal to the original image size. More specifically, as shown in Figure~\ref{fig:esa} (b) of the paper, this output is duplicated with size $W$ in the horizontal direction, where $W$ is the width of the input image. The expanded matrix is ESA matrix, $M^{h}_{esa}\in\mathbb{R}^{C\times H\times W}$. It should be noted that each row of $M^{h}_{esa}$ has the same element value, as shown in Figure~\ref{fig:esa} (b). Similarly, regarding the VESA module, the output of $f^{v}_{esa}$ of size $\mathbb{R}^{C\times 1\times W}$ is vertically expanded to ensure that the ESA matrix is $M^{v}_{esa}\in\mathbb{R}^{C\times H\times W}$, where $W$ is the width of the image. Therefore, as illustrated in Figure~\ref{fig:esa} (b), each column of $M^{v}_{esa}$ has the same value. The ESA matrix has a value between 0 and 1 owing to the sigmoid layer of $f_{esa}$ and highlights a part of the predicted probability map via the element-wise product between the predicted probability map and ESA matrix. If the predicted probability map is $P_{pred}\in\mathbb{R}^{\left(C+1\right)\times H\times W}$, the weighted probability map $E_{pred}$ is formulated as $E_{pred}=P_{pred}\odot M^{h}_{esa}$ for the HESA module and $E_{pred}=P_{pred}\odot M^{v}_{esa}$ for the VESA moduel, where the operator $\odot$ describes an a element-wise product. The reason that the number of channels in $P_{pred}$ is $C+1$ is that $C$ lane classes and one background class are included in the dataset. Therefore, element-wise product is performed only on lane channels except for a background channel, and the size of $E_{pred}$ is $C\times H\times W$.

The most important role of the ESA module is extracting lane confidence. Figure~\ref{fig:conf} presents the predicted probability map of the model and output of the ESA encoder $f^{h}_{esa}$. The colors in the graph match the colors of the lane. The output of $f^{h}_{esa}$ is identical to the height of the image. However, in Figure~\ref{fig:conf}, only the location in which the lane exists is presented as a graph. If there is no occlusion on the road as shown in the first figure in Figure~\ref{fig:conf}, the output of $f^{h}_{esa}$ is overall high. If occlusion occurs, such as the blue and yellow lanes in the second figure, the measured $f^{h}_{esa}$ value of the occluded area is small. This is how the ESA module measures the confidence of the lane. If the visual cues for the lane are abundant, the lane confidence at the location increases, and a great weight is output. Conversely, if there are few visual cues, the lane confidence decreases and a small weight is output.

\subsection{Network Architecture}
Our network architecture is illustrated in Figure~\ref{fig:network}. Our neural network starts with the baseline model, which consists of encoder and decoder. In this paper, since inference time is an important factor in lane detection, lightweight baseline models such as ResNet-18 \cite{he2016deep}, ResNet-34 \cite{he2016deep}, and ERFNet \cite{romera2017erfnet} are used. Inspired by the works~\cite{hou2019learning, liu2020lane}, we add the existence branch to the baseline model. Existence branch is designed for datasets in which lanes are classified according to their relative position, such as TuSimple and CULane. In the case of BDD100K, existence branch is not used because we consider all lanes as one class. We extract a total of four feature maps from the baseline model encoder. These feature maps are resized and concatenated to become input to the ESA module. We will discuss in detail how the ESA module output, baseline model output, and ground truth labels interact with each other in Section~\ref{loss}.

\subsection{Objective Functions}
\label{loss}
\noindent
\textbf{Segmentation and existence loss.} First we reduce the difference between the predicted lane segmentation map $S_{pred}$ and the ground truth segmentation map $S_{gt}$. The segmentation loss $\mathcal{L}_{seg}$ is used as follows:

\begin{equation}
	\mathcal{L}_{seg} = \mathcal{L}_{CE}\left(S_{pred},S_{gt}\right),
	\label{eq:one}
\end{equation}

\noindent
where $\mathcal{L}_{CE}$ is the standard cross entropy loss. We apply cross entropy loss to $C$ lane classes and one background class. In addition, the existence loss is proposed for the TuSimple and CULane datasets because lanes are classified by their relative positions. The existence loss $\mathcal{L}_{exist}$ is formulated as follows:

\begin{equation}
	\mathcal{L}_{exist} = \mathcal{L}_{BCE}\left(l_{pred},l_{gt}\right),
	\label{eq:two}
\end{equation}

\noindent
where $\mathcal{L}_{BCE}$ is the binary cross entropy loss, $l_{gt}$ is a lane existence label, and $l_{pred}$ is an output of the lane existence branch.

\begin{figure}
	\setlength{\belowcaptionskip}{-24pt}
	\begin{center}
		\subfloat
		{\includegraphics[scale=0.8]{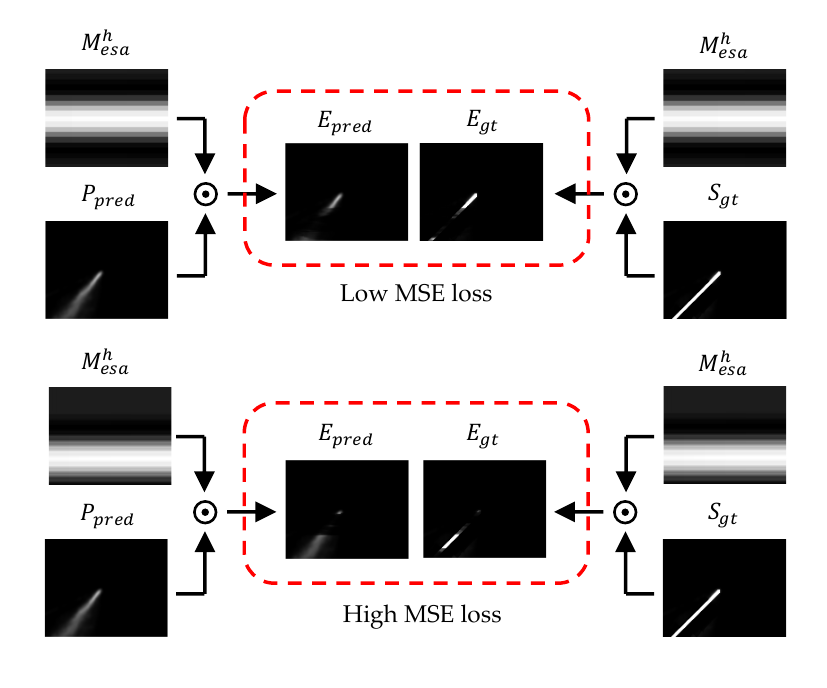}}\,
		\\[-1.5ex]
		\caption{Comparison of low (top) and high (bottom) loss. The mean square error is determined according to the location in which the ESA matrix is active.}
		\label{fig:mse}
	\end{center}
\end{figure}

\noindent
\textbf{ESA loss.} The ESA module aims to predict the confidence of the lane by recognizing occlusion with global contextual information. However, creating an annotation for the location information of the occlusion is time-consuming and expensive, and the consistency of the annotation cannot be guaranteed. Therefore, our module learns the occlusion location without additional annotations by reducing the mean square error between the weighted probability map $E_{pred}$ and the weighted ground truth segmentation map $E_{gt}$. Figure~\ref{fig:mse} presents this process. 

The predicted probability map of the lane is $P_{pred}=\Phi\left(S_{pred}\right)$, where $\Phi(.)$ is the softmax operator. In addition, the ESA loss $\mathcal{L}_{esa}$ is formulated as follows:

\begin{equation}
	\begin{split}
		\mathcal{L}_{esa} &=\mathcal{L}_{MSE}\left(E_{pred},E_{gt}\right)\\
		&+\lambda\left|\Psi\left(E_{pred}\right)-\Upsilon\Psi\left(S_{gt}\right)\right|,
		\label{eq:three}
	\end{split}
\end{equation}

\begin{table*}[t]
	\label{culane_table}
	\centering
	\footnotesize{
		\begin{tabular}{c|c|c||c|c|c|c|c}
			\hline
			Category & \textbf{R-34-H\&VESA} & \textbf{ERFNet-H\&VESA} & SCNN~\cite{pan2017spatial} & ENet-SAD~\cite{hou2019learning} & R-34-Ultra~\cite{qin2020ultra} & ERFNet~\cite{romera2017erfnet} & ERFNet-E2E~\cite{yoo2020end} \\
			\hline \hline
			Normal & 90.5 & \textbf{92.0} & 90.6 & 90.1 & 90.7 & 91.5 & 91.0 \\
			%\hline
			Crowded & 68.3 & \textbf{73.1} & 69.7 & 68.8 & 70.2 & 71.6 & \textbf{73.1} \\
			%\hline
			Night & 65.7 & \textbf{69.5} & 66.1 & 66.0 & 66.7 & 67.1 & 67.9 \\
			%\hline
			No line & 42.2 & 45.8 & 43.4 & 41.6 & 44.4 & 45.1 & \textbf{46.6} \\
			%\hline
			Shadow & 65.1 & \textbf{75.1} & 66.9 & 65.9 & 69.3 & 71.3 & 74.1 \\
			%\hline
			Arrow & 85.4 & \textbf{88.1} & 84.1 & 84.0 & 85.7 & 87.2 & 85.8 \\
			%\hline
			Dazzle light & 59.8 & 63.1 & 58.5 & 60.2 & 59.5 & \textbf{66.0} & 64.5 \\
			%\hline
			Curve & 61.5 & 68.8 & 64.4 & 65.7 & 69.5 & 66.3 & \textbf{71.9} \\
			%\hline
			Crossroad & \textbf{1675} & 2001 & 1990 & 1998 & 2037 & 2199 & 2022 \\
			%\hline
			Total & 70.9 & \textbf{74.2} & 71.6 & 70.8 & 72.3 & 73.1 & 74.0 \\
			\hline \hline
			Runtime (ms) &  \textbf{4.7} & 8.1 & 133.5 & 13.4 & 5.7 & 8.1 & \--  \\
			Parameter (M) & 2.44 & 2.68 & 20.72 & \textbf{0.98} & \-- & 2.68 & \--  \\
			\hline
		\end{tabular}
	}
	\caption{Comparison of F1-measures and runtimes for CULane test set. Only the FP is shown for crossroad. "R-" denotes "ResNet" and same abbreviation is used in the following sections.}
	\label{table:one}
	\vspace{-2ex}
\end{table*}

\noindent
where the ESA matrix is $M_{esa}$, the weighted probability map $E_{pred}=P_{pred}\odot M_{esa}$, the weighted ground truth map $E_{gt}=S_{gt}\odot M_{esa}$, and $\mathcal{L}_{MSE}$ is the mean square error loss. Moreover, the operator $\Psi(.)$ calculates the average of all values of the feature map, and $\lambda$ is a regularization coefficient. The coefficient $\Upsilon$ has an important effect on the performance of the model, and it determines the proportion of the weighted lane area. The first term on the right-hand side of Equation (\ref{eq:three}) is visualized in Figure~\ref{fig:mse}. In general, the lane probability map is blurred in areas with sparse supervisory signals. As shown in Figure~\ref{fig:mse}, if a large weight is given to the accurately predicted region in the probability map, the mean square error is small. Conversely, when a large weight is given to an uncertainly predicted area, the mean square error is large. This is how to predict the confidence of the lane without additional annotations.

In fact, if only the mean square error loss is used as the ESA loss, the ESA module outputs are all zeros in the training. To solve this problem, a second term is added as a regularizer to the right-hand side of Equation (\ref{eq:three}). This regularization term keeps the average pixel value of the weighted probability map equal to a certain percentage of the average pixel value of the ground truth map. This ratio is determined by $\Upsilon$, which has a value between 0 and 1.

It should be noted that although one ESA module is an HESA or a VESA module, both modules can be simultaneously attached to the model. In that case, the ESA loss is $\mathcal{L}_{esa}=\mathcal{L}^{h}_{esa}+\mathcal{L}^{v}_{esa}$, where $\mathcal{L}^{h}_{esa}$ is the ESA loss of the HESA module, and $\mathcal{L}^{v}_{esa}$ is the ESA loss of the VESA module. Finally, the above losses are combined to form the final objective function:

\begin{equation}
	\mathcal{L}=\alpha\mathcal{L}_{seg}+\beta\mathcal{L}_{exist}+\gamma\mathcal{L}_{esa}.
	\label{eq:four}
\end{equation}

\noindent
The parameters $\alpha$, $\beta$ and $\gamma$ balance the segmentation loss, existence loss, and ESA loss of the final objective function.

\section{Experiments}
\noindent
\textbf{Datasets.} We use three popular lane detection datasets TuSimple \cite{tusimple}, CULane \cite{pan2017spatial}, and BDD100K \cite{yu2018bdd100k} for our experiments. TuSimple datasets consist of images of highways with constant illumination and good weather, and are relatively simple datasets because the roads are not congested. Therefore, various algorithms \cite{pan2017spatial, neven2018towards, ghafoorian2018gan, hou2019learning, jung2020towards} have been tested on TuSimple datasets since before 2018. CULane is a very challenging dataset that contains crowded environments with city roads and highways with varying lighting conditions. The CULane dataset and TuSimple dataset are officially labeled with up to four lanes and one background excluding lanes. These datasets focus on the detection of four lane markings, which are paid most attention to in real applications. The BDD100K dataset also consists of images captured under various lighting and weather conditions. In addition, the largest number of lanes among the three datasets is labeled. However, because the number of lanes is large and inconsistent, we detect lanes without distinguishing instances of lanes.

\noindent
\textbf{Evaluation metrics.} 

\noindent
\textit{1) TuSimple.} In accordance with \cite{tusimple}, the accuracy is expressed as $Accuracy={N_{pred}\over N_{gt}}$, where $N_{pred}$ is the number of predicted correct lane points and $N_{gt}$ is the number of ground truth lane points. Furthermore, false positives (FP) and false negatives (FN) in the evaluation index.

\noindent
\textit{2) CULane.} In accordance with the evaluation metric in \cite{pan2017spatial}, each lane is considered 30 pixel thick, and the intersection-over-union (IoU) between the ground truth and prediction is calculated. Predictions with IoUs greater than 0.5 are considered true positives (TP). In addition, the F1-measure is used as an evaluation metric and is defined as follows:

\begin{equation}
	F_1={{2\times Precision\times Recall} \over {Precision+Recall}},
	\label{eq:five}
\end{equation}

\noindent
where $P recision={{TP} \over {TP+FP}}$, $Recall={{TP} \over {TP+FN}}$.

\noindent
\textit{3) BDD100K.} In general, since there are more than 8 lanes in an image, following~\cite{hou2019learning}, we determine the pixel accuracy and IoU of the lane as evaluation metrics.

We used different evaluation method for fair comparisons with previous studies. We evaluated with the same method as~\cite{ghafoorian2018gan, hou2019learning, neven2018towards, pan2017spatial} for TuSimple and~\cite{hou2019learning, pan2017spatial, qin2020ultra, yoo2020end} for CULane.

\begin{table}[!t]
	\centering
	\scriptsize{
		\begin{tabular}{c|c|c|c|c}
			\hline
			Algorithm & Accuracy & FP & FN & Runtime (ms)\\
			\hline \hline
			ResNet-18~\cite{he2016deep} & 92.69\% & 0.0948 & 0.0822 & \textbf{3.4} \\
			ResNet-34~\cite{he2016deep} & 92.84\% & 0.0918 & 0.0796 & 4.7 \\
			LaneNet~\cite{neven2018towards} & 96.38\% & 0.0780 & 0.0244 & 19.0 \\
			EL-GAN~\cite{ghafoorian2018gan} & 96.39\% & 0.0412 & 0.0336 & \-- \\
			ENet-SAD~\cite{hou2019learning} & \textbf{96.64\%} & 0.0602 & 0.0205 & 13.4 \\
			SCNN~\cite{pan2017spatial} & 96.53\% & 0.0617 & \textbf{0.0180} & 133.5 \\
			\hline \hline
			\textbf{R-18-H\&VESA} & 95.70\% & 0.0588 & 0.0622 & \textbf{3.4} \\
			\textbf{R-34-H\&VESA} & 95.83\% & 0.0587 & 0.0599 & 4.7 \\
			\textbf{ERFNet-HESA} & 96.01\% & \textbf{0.0329} & 0.0458 & 8.1 \\
			\textbf{ERFNet-VESA} & 95.94\% & 0.0340 & 0.0451 & 8.1 \\
			\textbf{ERFNet-H\&VESA} & 96.12\% & 0.0331 & 0.0450 & 8.1 \\
			\hline
			% \textbf{ResNet-18 (ours)} & 0.9269 & 0.0732 & 0.1102 \\
			% \hline
		\end{tabular}
	}
	\caption{Comparison of performance results of different algorithms applied to TuSimple test set.}
	\label{table:two}
\end{table}

\begin{table}[!t]
	\centering
	\footnotesize{
		\begin{tabular}{c|c|c|c}
			\hline
			Algorithm & Accuracy & IoU & Runtime (ms)\\
			\hline \hline
			ResNet-18~\cite{he2016deep} & 54.59\% & 44.63 & \textbf{2.7} \\
			ResNet-34~\cite{he2016deep} & 56.62\% & 46.00 & 4.1 \\
			ERFNet~\cite{romera2017erfnet} & 55.36\% & 47.04 & 7.3 \\
			ENet-SAD~\cite{hou2019learning} & 57.01\% & 47.72 & 12.1 \\
			SCNN~\cite{pan2017spatial} & 56.83\% & 47.34 & 123.6 \\
			\hline \hline
			\textbf{R-18-H\&VESA} & 57.03\% & 46.50 & \textbf{2.7} \\
			\textbf{R-34-H\&VESA} & 59.93\% & 49.51 & 4.1 \\
			\textbf{ERFNet-HESA} & 57.47\% & 48.97 & 7.3 \\
			\textbf{ERFNet-VESA} & 57.51\% & 48.24 & 7.3 \\
			\textbf{ERFNet-H\&VESA} & \textbf{60.24\%} & \textbf{51.77} & 7.3 \\
			\hline
		\end{tabular}
	}
	\caption{Comparison of results for BDD100K test set.}
	\label{table:three}
	\vspace{-2ex}
\end{table}

\begin{figure*}
	\setlength{\belowcaptionskip}{-10pt}
	\begin{center}
		\includegraphics[width=0.9\linewidth]{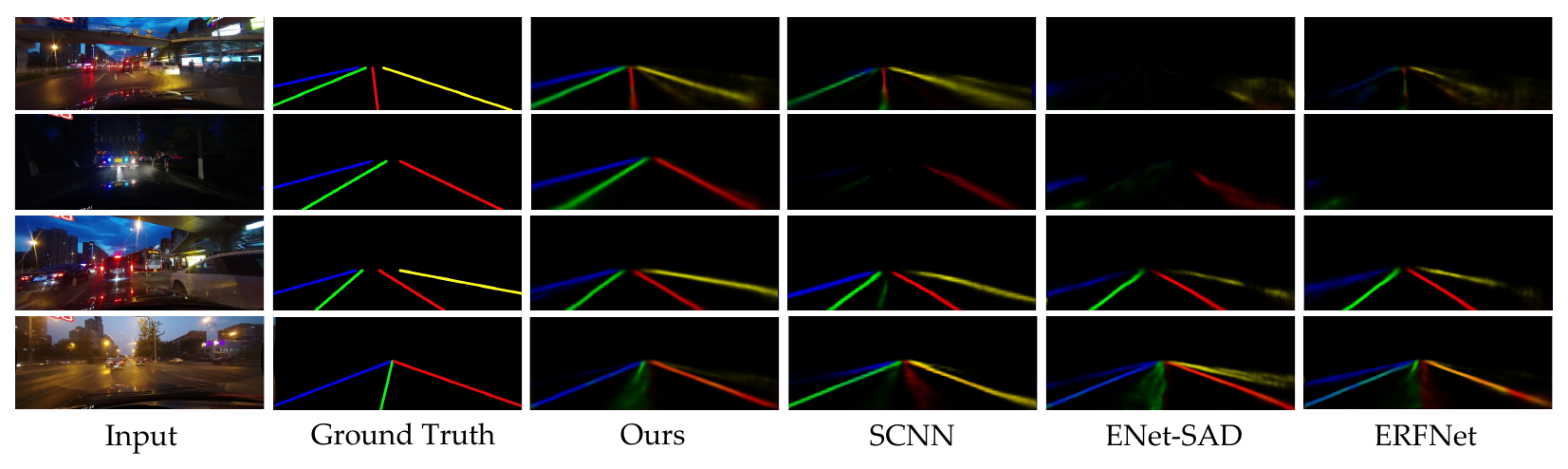}
		\\[-2ex]
		\caption{Comparison of the output probability maps of different algorithms applied to CULane test set. The third column is the result of the proposed ERFNet-HESA. The probability maps of the four lane classes are displayed in blue, green, red, and yellow, respectively.}
		\label{fig:culane}
	\end{center}
	\vspace{-2ex}
\end{figure*}

\noindent
\textbf{Implementation details.} Following \cite{pan2017spatial, hou2019learning}, we resize the images of TuSimple, CULane, and BDD100K to $368\times640$, $288\times800$, and $360\times640$, respectively. The original BDD100K images label one lane with two lines. Because this labeling method is difficult to learn, so we drew new 8 pixel thick ground truth labels that pass through the center of the lane. The new ground truth labels are applied equally to both train and test sets. Moreover, SGD~\cite{bottou2010large} is used as the optimizer, and the initial learning rate and batch size are set to 0.1 and 12, respectively. The loss balance coefficients $\alpha$, $\beta$, and $\gamma$ in Equation (\ref{eq:four}) are set to 1, 0.1, and 50, respectively. The regularization coefficient $\lambda$ in Equation (\ref{eq:three}) is 1. It is experimentally verified whether the value of the coefficient $\Upsilon$ in Equation (\ref{eq:three}) has a significant effect on the performance of the model. In CULane and BDD100K, the optimal $\Upsilon$ value is set to 0.8, and TuSimple is set to 0.9. The effect of $\Upsilon$ on the performance is discussed in detail in Section~\ref{ablation}. Because the BDD100K experiment regards all lanes as one class, the output of the original segmentation branch is replaced with a binary segmentation map. In addition, the lane existence branch is removed for the evaluation. All models are trained and tested with PyTorch and the Nvidia RTX 2080Ti GPU.

\subsection{Results}
\label{result}
Tables~\ref{table:one}-\ref{table:three} compare the performance results of the proposed method and previously presented state-of-the-art algorithms for CULane, TuSimple, and BDD100K datasets. The proposed method is evaluated with the baseline models ResNet-18 \cite{he2016deep}, ResNet-34 \cite{he2016deep}, and ERFNet \cite{romera2017erfnet}, and each model is combined with either an HESA or a VESA. Moreover, the use of both HESA and VESA modules is denoted as “H\&VESA”. The effects of using both modules simultaneously are presented in Section~\ref{ablation}.

\begin{figure}
	\setlength{\belowcaptionskip}{-24pt}
	\begin{center}
		\includegraphics[width=\linewidth]{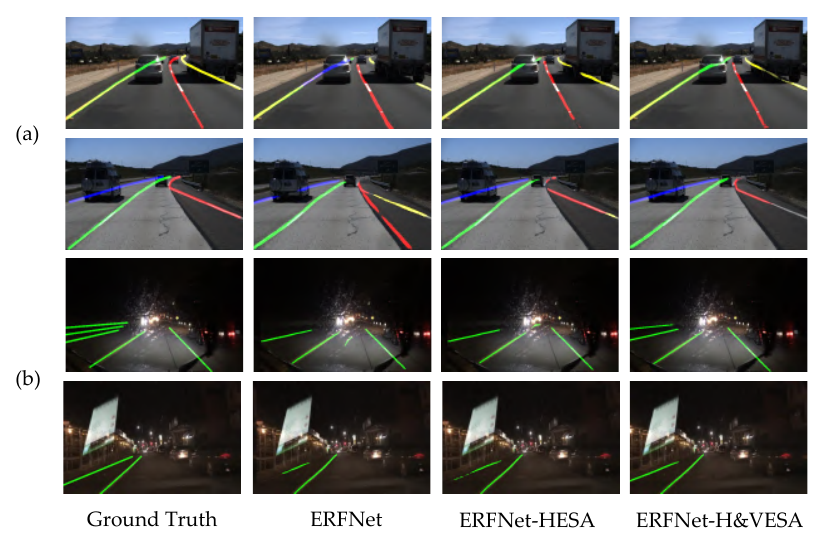}
		\caption{Performance of different algorithms for (a) TuSimple and (b) BDD100K test sets.}
		\label{fig:bdd}
	\end{center}
\end{figure}

\begin{table*}[!t]
	\centering
	\footnotesize{
		\begin{tabular}{c|c c|c|c|c|c|c|c}
			\hline
			\multirow{2}*{Baseline} & \multirow{2}*{HESA} & \multirow{2}*{VESA} & \multicolumn{3}{|c|}{TuSimple} & CULane & \multicolumn{2}{|c}{BDD100K} \\
			\cline{4-9}
			~& & & Accuracy & FP & FN & F1 (total) & Accuracy & IoU \\
			\hline \hline
			\multirow{4}*{ResNet-18~\cite{he2016deep}} & & & 92.69\% & 0.0948 & 0.0822 & 67.8 & 54.59\% & 44.63 \\
			& \ding{51} & & \textbf{95.73\%} & 0.0590 & 0.0643 & 70.4 & 56.68\% & 46.11 \\
			& & \ding{51} & 95.70\% & 0.0598 & \textbf{0.0615} & 70.3 & 56.70\% & 46.08 \\
			& \ding{51} & \ding{51} & 95.70\% & \textbf{0.0588} & 0.0622 & \textbf{70.7} & \textbf{57.03\%} & \textbf{46.50} \\
			\hline
			\multirow{4}*{ResNet-34~\cite{he2016deep}} & & & 92.84\% & 0.0918 & 0.0796 & 68.4 & 56.62\% & 46.00 \\
			& \ding{51} & & 95.68\% & 0.0584 & 0.0634 & 70.7 & 58.36\% & 47.29 \\
			& & \ding{51} & 95.70\% & \textbf{0.0533} & 0.0681 & 70.7 & 58.11\% & 47.30 \\
			& \ding{51} & \ding{51} & \textbf{95.83\%} & 0.0587 & \textbf{0.0599} & \textbf{70.9} & \textbf{59.93\%} & \textbf{49.51} \\
			\hline
			\multirow{4}*{ERFNet~\cite{romera2017erfnet}} & & & 94.90\% & 0.0379 & 0.0563 & 73.1 & 55.36\% & 47.04 \\
			& \ding{51} & & 96.01\% & \textbf{0.0329} & 0.0458 & \textbf{74.2} & 57.47\% & 48.97 \\
			& & \ding{51} & 95.94\% & 0.0340 & 0.0451 & 74.1 & 57.51\% & 48.24 \\
			& \ding{51} & \ding{51} & \textbf{96.12\%} & 0.0331 & \textbf{0.0450} & \textbf{74.2} & \textbf{60.24\%} & \textbf{51.77} \\
			\hline
		\end{tabular}
	}
	\caption{Performance comparison of various combinations of HESA and VESA modules with TuSimple, CULane, and BDD100K test sets}
	\label{table:four}
	\vspace{-3ex}
\end{table*}

The combination of the baseline model ERFNet and ESA module outdoes the performance of the ERFNet and achieves state-of-the-art performance for CULane and BDD100K. In particular, ERFNet-H\&VESA provides significant performance gains for almost all driving scenarios in the CULane dataset compared to ERFNet. However, the runtime and number of parameters remain unchanged. In addition, ERFNet-H\&VESA surpasses the existing methods by achieving an F1-measure of 69.5 in the challenging low-light environment in the lane detection with the CULane dataset. It has a fast runtime similar to those of the previous state-of-the-art methods in Table~\ref{table:one}. Thus, the proposed method is much more efficient than the previously proposed methods. As shown in Table~\ref{table:three}, compared to ERFNet, ERFNet-HESA increases accuracy from 55.36\% to 57.47\% with the BDD100K dataset. In addition, ERFNet-H\&VESA achieves the highest accuracy of 60.24\%. These results show that the HESA and VESA modules work complementarily. The regarding details are covered in Section~\ref{ablation}. The results of the TuSimple dataset in Table~\ref{table:two} show the effect of the ESA module, but it does not achieve the highest performance. The TuSimple dataset contains images of highways with bright light, and generally less occlusion. Because the ESA module extracts global contextual information by predicting the occluded location, our method is less effective for datasets with less occlusion. Unlike our method, ENet-SAD~\cite{hou2019learning} provides additional supervision signals to lane areas and SCNN~\cite{pan2017spatial} applies a message passing mechanism between adjacent pixels in visible lanes. EL-GAN~\cite{ghafoorian2018gan} uses adversarial learning to capture sparse supervision from visible lanes or sharpen blurry lanes. Therefore, these methods are effective in datasets with less occlusion. 

We provide qualitative results of our algorithm for various driving scenarios in three benchmarks. In particular, the first and second rows of Figure~\ref{fig:culane} show that our method can detect sharp lanes even under extreme lighting conditions and in situations in which the lanes are barely visible owing to other vehicles. Figure~\ref{fig:bdd} (a) shows that the ESA module can connect the lanes occluded by vehicles without interruption. According to Figure~\ref{fig:bdd} (b), the approach achieves more accurate lane detection in low-light environments. Thus, compared to the baseline model, the ESA module can improve performance in challenging driving scenarios with extreme occlusion and lighting conditions.

\subsection{Ablation Study}
\label{ablation}

\noindent
\textbf{Combination of HESA and VESA.} Table~\ref{table:four} summarizes the performance characteristics of different combinations of HESA and VESA. The following observations can be made. (1) The performance characteristics of the HESA and VESA modules are similar. (2) In general, the performance of H\&VESA with HESA and VESA modules applied simultaneously is better. In addition, H\&VESA results in a remarkable performance improvement for BDD100K. The reason why the HESA and VESA modules lead to similar performance characteristics is that the predicted direction of the lane confidence is not important for extracting the low-level occlusion location because the lane has a simple geometric shape. Because the HESA and VESA modules complement each other to extract more abundant global contextual information, it is not surprising that H\&VESA generally achieves the highest performance. Therefore, global contextual information is more important for the BDD100K dataset, which includes many lanes.

\begin{figure}
	\setlength{\belowcaptionskip}{-24pt}
	\begin{center}
		\includegraphics[width=0.8\linewidth]{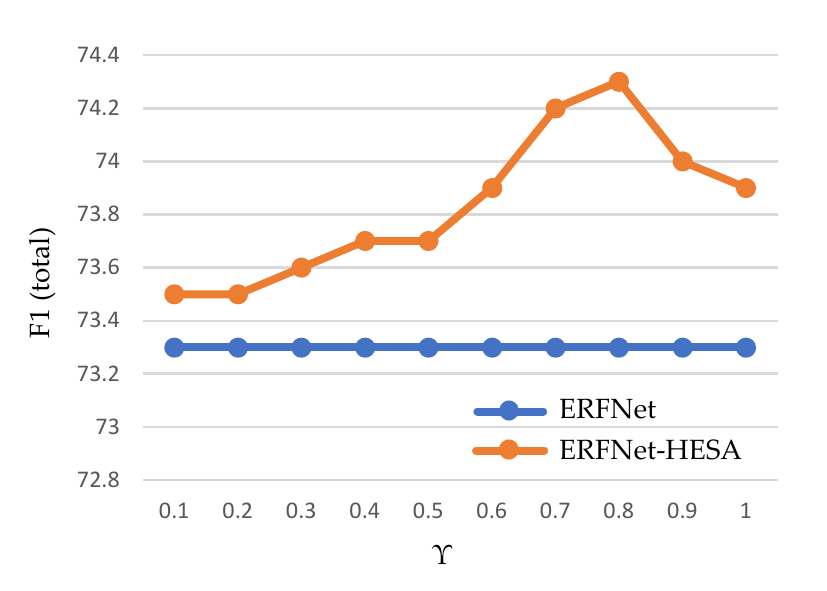}
		\vspace{-3ex}
		\caption{Comparison of performance characteristics with respect to $\Upsilon$ for the CULane validation set.}
		\label{fig:graph}
	\end{center}
\end{figure}

\noindent
\textbf{Value of $\Upsilon$.} Figure~\ref{fig:graph} compares the total F1-score of the CULane validation set with respect to $\Upsilon$ in Equation~(\ref{eq:three}). As shown in Figure~\ref{fig:graph}, the model shows the best performance at $\Upsilon=0.8$ in ERFNet-HESA. It is important to find a suitable $\Upsilon$ value because it determines the ratio of occluded and normal areas. When the $\Upsilon$ is small (\ie, when the predicted occlusion area is wide), the sensitivity to occlusion decreases, which makes it difficult to determine the occluded location accurately. Conversely, when $\Upsilon$ is large, the detected occlusion area becomes narrow, which makes it difficult for the network to reinforce learning for the entire occluded area.

\section{Conclusion}
This paper proposes ESA module, a novel self-attention module for robust lane detection in occluded and low-light environments. The ESA module extracts global contextual information by predicting the confidence of the lane. The performance of the model is evaluated on the datasets containing a variety of challenging driving scenarios. According to the results, our method outperforms previous methods. We confirm the effectiveness of the ESA module in various comparative experiments and demonstrate that our method is robust in challenging driving scenarios.

\noindent\footnotesize\textbf{Acknowledgement.} This research was supported by Multi-Ministry Collaborative R\&D Program(R\&D program for complex cognitive technology) through the National Research Foundation of Korea(NRF) funded by MSIT, MOTIE, KNPA(NRF-2018M3E3A1057289).

{\small
\bibliographystyle{ieee_fullname}
\bibliography{egbib}
}

\end{document}